%% file: neurips_2019.tex
\documentclass{article}

\usepackage[preprint]{neurips_2019}

\usepackage[utf8]{inputenc} 
\usepackage[T1]{fontenc}    
\usepackage{url}            
\usepackage{booktabs}       
\usepackage{amsfonts}       
\usepackage{nicefrac}       
\usepackage{microtype}      

\usepackage{graphicx}
\usepackage{subfigure}
\usepackage{multirow}
\usepackage{natbib}
\usepackage{threeparttable}
\usepackage{makecell}
\usepackage{booktabs} 
\usepackage{float} 

\usepackage{amsfonts,amsmath,amssymb,amsthm,mathtools} 
\usepackage{wrapfig}  
\usepackage{mdwlist}  
\usepackage{sidecap}  
\usepackage{bbm}
\usepackage{bm,amsbsy} 

\usepackage{caption}
\usepackage{xcolor}
\usepackage{algorithm,algorithmic}

\usepackage{hyperref}       

\title{Neuron ranking -- an informed way to condense \\ convolutional neural networks architecture }

\author{%
 Kamil Adamczewski \\
  Max Planck Institute for Intelligent Systems\\
  \texttt{kamil.m.adamczewski@gmail.com} \\
   \AND
   Mijung Park \\
   Max Planck Institute for Intelligent Systems \\
   \texttt{mijung.park@tuebingen.mpg.de} \\
}

\input{notations}

\begin{document}

\maketitle

\begin{abstract}

Convolutional neural networks (CNNs) in recent years have made a dramatic impact in science, technology and industry, yet the theoretical mechanism of CNN architecture design remains surprisingly vague. The CNN neurons, including its distinctive element, convolutional filters, are known to be learnable features, yet their individual role in producing the output is rather unclear. The thesis of this work is that not all neurons are equally important and some of them contain more useful information to perform a given task . Consequently, we quantify the significance of each filter and rank its importance in describing input to produce the desired output. This work presents two different methods: (1) a game theoretical approach based on \textit{Shapley value} which computes the marginal contribution of each filter; and (2) a probabilistic approach based on what-we-call, the \textit{importance switch} using variational inference. Strikingly, these two vastly different methods produce similar experimental results, confirming the general theory that some of the filters are inherently more important that the others. The learned ranks can be readily useable for network compression and interpretability. 

\end{abstract}


\section{Introduction}

Neural network have achieved state-of-the art results in various cognition tasks, including image and speech recognition, machine translation, reinforcement learning \citep{1211479, mnih2013playing, GU2018354}. Many of these applications involved convolutional neural networks which excel in particular in the vision tasks due to its ability to capture visual by means of convolution filters. Although the effectiveness of convolutional networks is unquestionable, the details of the architecture design and what particularly makes neural network work in detail remain highly uncertain. The experimental results roughly confirm that the accuracy of the network and representational capacity is correlated with the depth of the network, however the exact number of layers is unknown \citep{simonyan2014very, he2016deep, montufar2014number}. Interestingly, the deeper architecture also become wider, although the link between width and network expressivity is questionable \citep{poole2016exponential} and the choice of the number of neurons is rather discretionary. As a result the discussion about the network architecture often revolves around the numbers of filters and layers and their relative positioning, putting aside the conversation about the quality of the information that it contains. 

The increasing size of the network architectures have faced scrutiny that made claims that the networks are overparametrized raising two main concerns: heavy computational load and potential overfitting \citep{2017arXiv171201312L}. In response to the need to build networks that are smaller yet accurate, a stream of research attempted to remove redundant units, compress the networks and design lighter architectures \citep{iandola2016squeezenet, ullrich2017soft}. A widespread approach to network reduction has been removing weights that are small or even close to zero \citep{han2015deep}. This line of research implicitly discerns that nodes with larger weights are more significant for learning task than the small weights. As a result, broadly speaking, this approach divides features between those that are useful which are kept and those which are insignificant and therefore discarded, forming a sort of binary approach. 

In this work, we would like to form an explicit theory that states that the units in the network (both convolutional filters and nodes in fully connected layers) are not equally important when it comes to performing an inference task. The corollary of this thesis is that CNNs learn features in a discriminative way so that some of them carry more significance than others, and the knowledge about the input is not uniformly distributed among the CNN features. This theory is in line of research that adding more filters does not make the network more expressive since learning relevant information to the network has already been addressed by other filters.

Given the proposed theory, we would like to make a step forward in gaining insight what the CNN learns and propose to extend the binary approach to form a quantifiable ranking of features. In other words, we attempt to estimate the importance of each feature compared to the others with particular focus on convolutional filters, which may be visualized. We introduce a theoretical framework to quantify how important each feature is through proposing a feature ranking method based on two different approaches. 

The first approach derives from the game theoretical concept of \textit{Shapley value} \citep{shapley1953}, which assesses the marginal contribution of an individual in a group of agents. Assuming that an important feature allows for task generalization, the feature importance further translates into finding features that carry most information and usefulness in the prediction task that lead to achieving higher accuracy. The second method takes a probabilistic approach and introduces additional learnable parameters, which we call \textit{importance switches}, that take real values and are trained by means of variational inference to give more weight to the important features. 
The experimental results produce the most notable realization of this work that both methods produce very similar ranking of features, which works in favor of the proposed theory that the some features are inherently more significant regardless of the choice of rank method. 

The theoretical underpinnings of the feature rankings have further direct practical implications we explore. Firstly, the knowledge of the ranking allows to know which features directly impact the score of our method and consequently a more informed way of building an effective model. Thus, we are able to build a network around the the relevant features and discard the less relevant ones, effectively compressing the network achieving state-of-the-art results.
Secondly and perhaps more significantly, the feature ranking of convolutional features provides more interpretable information about the network and places meaning on particular features in the context of a given task, thus casting light on the black box models. To achieve human interpretability, we visualize the most significant features which significantly show the significance of repeated and complementary features.

The rest of the paper is organized as follows. We start by providing an overview of related work in \secref{relatedwork}. Then, we describe the game theoretic neuron ranking method in \secref{Shapley}, and our probabilistic approach in \secref{Switches}.
Finally in \secref{Experiments} we empirically show that our method improves compression ability over existing methods and provides interpretable ranking (i.e., importance) of learned neurons.   





\section{Related works}\label{sec:relatedwork}

In early years of convolutions neural network development the network were limited to a few layers \citep{lecun1998gradient}. With a line of new work in recent years, the architectures have become deeper and wider \citep{krizhevsky2012imagenet, szegedy2015going}. The emergence of GPU implementability and regularization algorithms \citep{srivastava2014dropout, ioffe2015batch} has  allowed to use large architectures which train and generalize well. Nevertheless, the trend towards building larger neural networks ironically opposed the research about the nature, interpretability and knowledge extraction from within the neural network models, which we are interested in this work. 
Therefore, we will compare our method to existing ones in terms of compression ability and interpretability, and then frame the idea in terms of feature selection and feature ranking in the next section.

\paragraph{Compression.} 
%
The early work on compression largely focused on non-Bayesian approaches (e.g.,\citep{hassibi1993second}) and  mostly centered around non-structured pruning methods, e.g., removing single weights from the architectures of CNNs \citep{han2015deep}. Then, hardware-oriented structured pruning techniques made more practical speed-ups (e.g., \citep{srinivas2015data,li2016pruning, wen2016learning,Lebedev_2016, zhou2016less}). This work is in line with structured compression line of work. We believe that although non-structured compression may be interesting theoretically, the practical implications of structured compression are far more relevant.
More recently Bayesian methods using the network weights' uncertainties 
have achieved impressive compression rates, e.g., using sparsity inducing prior on scale parameter \citep{molchanov2017variational}, using Gaussian mixture priors \citep{ullrich2017soft}, and using the grouping of weights through a group Horseshoe prior \citep{louizos2017bayesian}, among many. 
However, unlike our method, none of these methods prune neurons based on the learned importance of the filters.

\paragraph{Interpretability.} 
Broadly speaking, there are three main lines of work done for intepretability of CNNs. The first line of work, the early and used-to-be very popular work, focused on  visualization of neurons in CNNs to understand how information propagates within the network \citep{ZeilerF13, Simonyan14a, YosinskiCNFL15}. 
Another line of work focused on probing trained CNNs to obtain local \& pixel level explanations either layer-wise or class-wise using gradient information of querying points \citep{SelvarajuDVCPB16, journal.pone.0130140, MontavonBBSM15}. 
Last line of work focused on mapping semantic concepts to latent representations of CNNs \citep{BauZKOT17}. 
Other somewhat related work for interpretability using Shapley value also exist (but not in the context of CNNs) \citep{NIPS2017_7062}. Compared to existing methods, our method provides a global view of learned features, in terms of their importance for a given task.  




%
%

%
%

While there are many papers on compressing the network, in our view there is much less discussion about what the smaller and slimmer models actually contain.  In order to identify the significance of each neuron's contribution in  producing a desired output, 
%
%
%
we introduce a new term, \textit{neuronal feature selection}, or \textit{neuron selection}, which is a rephrase of the feature selection in the context of CNNs. 
%
%
%
%
The broad architectures and increasing network sizes emphasize \textit{quantity} as a characteristic which allows to build a comprehensive neural network. However, in the aim of building smaller networks, the \textit{quality} of the network architectures matters more. In fact, as hinted in \citep{rathore2014comparative}, feature ranking via neuron ranking indeed produces better results in many tasks.  
%
In what follows, we introduce the two methods to perform neuronal feature selection.

\section{Game theoretical neuron ranking}\label{sec:Shapley}








Let $N$ be the number of agents, which in this work are CNN features (also referred as neurons or nodes). To be specific, let $N_l$ to be the number of neurons in a layer $l$ (in unambiguous cases, for clarity we omit the subscript). 

\subsection{Coalitional game theory}

A coalitional game is a game where utility is given to a group of players (in our case, nodes or neurons) instead of each agent individually. For every group of players, a coalitional game specifies the payoff the members receive as a group or a coalition. We define a \textit{coalition} of the neurons $N$ of a layer $L$ as a subset of neurons, $C \subseteq N$. To assess quantitatively the performance of a group of agents, each coalition is assigned to a real number, which is interpreted as a payoff that a coalition receives from being together. Mathematically, the value of a coalition is given by a \textit{characteristic function}, which assigns a real number to a set of nodes. Formally, a characteristic function $\nu$: $2^{N} \rightarrow \mathbb{R}$ maps each coalition (subset) $C \subseteq N$  to a real number $\nu(C)$. Therefore, a coalitional game is defined by a tuple $(N, \nu)$, where $N$ is a set of players and $\nu$ is a function that assigns payoffs to every coalition of $N$.

A critical component of a coalitional game is specifying the choice of characteristic function that is assigned to a given subset of features. In the case of CNN, the test metric is accuracy which assesses whether the (argmax of) the network output is the correct label averaged over the number of examples. As a result, we choose the accuracy on a validation set as the characteristic function,
that is, $\nu(C)=\text{acc}(C)$ and $\nu(N)=\text{acc}(N)$. 
The question now remains how to assess the importance of a single feature given the information about the payoffs for each subset of nodes. To this end, we employ the concept of Shapley value about the normative payoff of the total reward/cost, that is a division scheme that allows to distribute the total payoff uniquely and in a fair way.

\subsection{Shapley value}

Shapley proposes to evaluate each player by the marginal contribution that the player makes to every coalition averaged over all the coalitions.  The marginal contribution of an agent $n$ is the difference between the value of a coalition $C$ that contains $n$ and the coalition $C \setminus n$. For example, when a coalition has no members, i.e. is empty, and the neuron $n_1$ joins the coalition, the value of its marginal contribution is equal to the value of the one-member coalition as the value of the empty coalition is equal to 0, $\nu(\{ n_1 \})-\nu (\{ \emptyset\})=\nu(\{ n_1 \})$ where $\{n_1\}=C$. Subsequently, when another agent $n_2$ joins this coalition, its marginal contribution is equal to $\nu(\{n_1, n_2\})-\nu(\{n_1\})$. The process continues until all the nodes join the coalition. The coalition of all the nodes is called the \textit{grand coalition}.


The order of nodes, which builds subsequent coalitions to finally the grand coalition, can be represented as a permutation of nodes. The examples of different permutations are $n_1n_2 ... n_{N-1}n_N$ or $n_5n_3n_7 ... n_N ... n_2$. One permutation represents one way to form a coalition. For example, in the case of permutation $n_5n_3n_7 ... n_N ... n_2$, the agent $n_5$ creates the first non-empty coalition on it own, then two-element coalition $n_5n_3$ is formed, and so on. All the subsequent nodes join the coalition in the order given by the permutation. There are $N!$ permutations of $N$ nodes, meaning that there are $N!$ different ways to form a coalition. To compute the Shapley value of the node $n$, we add the marginal contributions of $n$ for each of the $N!$ permutations and divide the sum  by all the permutations. The Shapley value of $n$ is then the averaged marginal contribution of $n$.


Formally, let $\pi$ denote a permutation, $\pi(i)$ a place of the neuron $n_i$ in the permutation $\pi$, and $C_{\pi}(i)$ the coalition formed by the predecessors of $n_i$ such that $C_{\pi}(i)=\{n_j \in \pi : \pi(j) \textrm{ before } \pi(i)\}$. For example, in the permutation $n_5n_3n_7 ... n_N ... n_2$, $\pi(3)=n_7$ and $C_{\pi}(3)=\{n_5, n_3\}$. The Shapley value ($SV_i$) of the node $n_i$ is thus defined as follows:
\begin{equation}
\label{eq:permutations}
SV(n_i)=\sum_{\pi \in \Pi(N)} \frac{1}{|N|!}(\nu(C_{\pi}(i) \cup \{n_i\})-\nu(C_{\pi}(i)))
\end{equation}
This formula can also be written in a form that considers sets instead of permutations:
\begin{equation}
\label{eq:sets}
SV(n_i)=\sum_{C \subseteq N \setminus\{v_i\}} \frac{|C|!(|N|-|C|-1)!}{|N|!}(\nu(C \cup \{n_i\})-\nu(C_{\pi}(i)))
\end{equation}

\subsubsection{Practical considerations}




%
Shapley value is a mathematically rigorous division scheme and, strictly speaking, it has been proposed as the only measure that satisfies four normative criteria regarding the fair payoff distribution. These criteria are (1) efficiency where the total gain is distributed among the agents, (2) symmetry; if $i$ and $j$ are agents such that $\nu(C \cup {i})=\nu(C \cup j)$ for each coalition $C$ of $N$, then $SV(i)=SV(j)$, (3) null player payoff such that an agent who contributes nothing to every coalition obtains zero individual payoff and (4) linearity; $\nu(C)=\nu_1(C)+\nu_2(C)$ for every coalition implies $SV_{\nu_1}(i)+SV_{\nu_2}(i)=SV_{\nu}(i)$. Nevertheless, the choice of a characteristic function which satisfies these criteria is not feasible in case of our application due to the fact that we do not have control over the output of the model. As a result for example, the characteristic function may not be monotone which violates the first criterion. However, the payoff produced by the Shapley value, although may not be unique, is a valid cost division which, as shown in the experimental section, works well in practice.

 The second practical consideration is that computing the characteristic function for every subset is combinatorial in nature. In the similar fashion, computing the Shapley value also takes exponential time complexity.
 Although it is still possible to compute the complete characteristic function for smaller number of features, for the larger number of features, the task is computationally unfeasible. 
 We propose the following solutions to approximate the optimal solution and obtain a sensible ranking metric based on Shapley value.
 The first solution entails computing the Shapley value for the subsets no larger than arbitrary $k$. As a result we only compute the synergies that are no larger than $k$. Intuitively, we assume that that the larger the coalition, the less information is to be obtained from computing the large subsets.
 The second solution is based on sampling and sampling provides an unbiased estimate of the optimal result.Thus, we first sample the characteristic function and then sample the permutations needed for the computations of the Shapley value.

%
What comes next describes our proposal to improve the speed of computation for identifying the ranking by introducing an additional operation called \textit{importance switch} per layer to the existing neural network architecture, and learning the optimal switch using variational inference to infer the neuron ranking.

\section{Probabilistic feature ranking}\label{sec:Switches}

\subsection{ Importance switches}
To infer the neuron ranking in each layer, we propose to make a slight modification in the existing neural network architecture. We introduce a component, the \textit{importance switch}, denoted by $\vs_{l}$ for each layer $l$. Each importance switch is a probability vector of length $D_l$ (the output dimension of the $l$th layer) and $\sum_j^{D_l} \vs_{l,j} = 1$, where $\vs_{l,j}$ is the $j$th element of the vector. With this addition, we rewrite the forward pass under a deep neural network model, where the function $f(\mW_l, \vx_i)$ can be the convolution operation for CNNs or simple matrix multiplication for MLPs between the weights $\mW_l$ and the unit $\vx_i$,  
\begin{align}
    &\mbox{Pre-activation followed by a switch $\vs_l$: }  \vh_{l,i} = \vs_l \circ \left[f(\mW_l, \vx_i)\right], \\
    &\mbox{Input to the next layer after going through a nonlinearity $\sigma$: }  \vz_{l,i} = \sigma(\vh_{l,i}),
\end{align} 
where $\circ$ is an element-wise product. Introducing a switch operation between layers in a neural network model was also presented in \citep{2017arXiv171201312L}, although in their case, the switch is a binary random variable (called a gate).
The output probability under such networks with $L$ hidden layers for solving classification problems can be written as
\begin{align}
    P(\vy_i|\vx_i, \{\mW_{l}\}_{l=1}^{L+1}) = g \left(\mW_{L+1} \vz_{L,i}\right), \mbox{ where }  \; \vz_{L,i} &= \sigma(\vs_L \circ \left[f(\mW_L \vz_{L-1,i}) \right]).
\end{align} where $g$ is the \textit{softmax} operation. 

\subsection{Variational learning of importance switches}
A natural choice to model the distribution over the switch is the \textit{Dirichlet} distribution, which defines a probability distribution over a probability vector. 
We model each switch as a vector of independent Dirichlet distributed random variables
\begin{align}
    p(\vs_l) &= \mbox{Dir}(\vs_l|\valpha_0).
\end{align} 
When there is no prior knowledge, i.e.,  \textit{a priori} we don't know which feature would be more important for prediction, so we treat them all equally important features by setting the same value to each parameter, i.e., $\valpha_0 = \alpha_0*\vone_{D_l}$ where $\vone_{D_l}$ is a vector of ones of length $D_l$.  When we apply the same parameter to each dimension, this special case of Dirichlet distribution is called \textit{symmetric} Dirichlet distribution. In this case,
if we set $\alpha_0<1$ , this puts the probability mass toward a few components, resulting in only a few components that are non-zero, i.e., inducing sparse probability vector. If we set $\alpha_0>1$, all components become similar to each other.

We model the posterior over $\vs_l$ as the Dirichlet distribution as well but with \textit{asymmetric} form to learn a different probability on different elements of the switch (or neurons), using a set of variational parameters (the parameters for the posterior). We denote the variational parameters by $\vphi_l$, where each element of the vector can choose any values above 0. Our posterior distribution over the switch is, hence, defined by  
\begin{align}
    q_{\vphi_l}(\vs_l) &= \mbox{Dir}(\vs_l|\vphi_l).
\end{align} 
With this parametric form of prior and posterior, we optimize the variational parameters $\vphi_l$ over each layer's importance switch by maximizing the variational lower bound with freezing all the weights to the pre-trained values,
\begin{align}\label{eq:ELBO}
    \log p(\Dat) \geq \LL(\vphi_l) := \int q_{\vphi_l}(\vs_l) \log p(\Dat|\vs_l) d\vs_l - \mbox{D}_{kl}[ q(\vs_l|\vphi_l) || p(\vs_l |\valpha_0)].
\end{align} 
We do this variational learning for each layer's importance switch sequentially from the input layer to the last layer before the output layer.

Computing the gradient of \eqref{ELBO} with respect to $\vphi_l$ requires computing the gradients of the integral (the first term on RHS) and also the KL divergence term (the second term on RHS), as both depends on the value of $\vphi_l$. 
The KL divergence between two Dirichlet distributions is wrttien in closed form,
\begin{align}
    \mbox{D}_{kl}[ q(\vs_l|\vphi_l) || p(\vs_l |\valpha_0)] &= \log \Gamma (\sum_{j=1}^{D_l} \vphi_{l,j}) - \log \Gamma (D_l \alpha_{0})
   - \sum_{j=1}^{D_l} \log \Gamma(\vphi_{l,j}) + D_l \log \Gamma(\alpha_{0}) \nonumber \\
    & \quad +  \sum_{j=1}^{D_l} (\vphi_{l,j} - \alpha_{0}) \left[ \psi(\vphi_j) - \psi(\sum_{j=1}^{D_l}\vphi_{l,j} ) \right],
\end{align} where $\vphi_{l,j}$ denotes the $j$th element of vector $\vphi_{l}$, $\Gamma$ is the Gamma function and $\psi$ is the digamma function. Back-propagation of the gradients of this KL divergence w.r.t.\ $\vphi_l$ is tractable using standard deep learning tools such as Pytorch.  

Unlike the KL term, the first term is tricky. 
As described in \citep{NIPS2018_7326}, the usual reparameterization trick, i.e., replacing a probability distribution with an equivalent parameterization of it by using a deterministic and differentiable transformation of some fixed base distribution\footnote{For instance, a Normal distribution for $z$  with parameters of mean $\mu$ and variance $\sigma^2$ can be written equivalently as $z = \mu + \sigma \epsilon$ using a fixed base distribution $\epsilon \sim \Nrm(0,1)$.}, does not work. For instance, in an attempt to find a reparameterization, one could adopt the representation of a $k$-dimensional Dirichlet random variable as a weighted sum of Gamma random variables,
   $ \vs_{l,j} = y_j/(\sum_{j'=1}^K y_{j'}), $ where 
    $y_j \sim \mbox{Gam}(\vphi_{l,j},1) = {y_j^{(\vphi_{l,j}-1)}\exp(-y_j)}/{\Gamma(\vphi_{l,j})}$,
    for $\vs_l \sim \mbox{Dir}(\vs_l|\vphi_l)$,
where the shape parameter of Gamma is $\vphi_{l,j}$ and the scale parameter is $1$. However, this does not allow us to detach the randomness from the parameters as the parameter still appears in the Gamma distribution, hence one needs to sample from the posterior every time the variational parameters are updated, which is costly and time-consuming.
Existing methods suggest either explicitly or implicitly computing the gradients of the inverse CDF of the Gamma distribution during training to decrease the variance of the gradients (e.g., \citep{2015arXiv150901631K} and \citep{NIPS2018_7326} among many). The length of the importance switch we consider is mostly less than on the order of $100$s, in which case the variance of gradients does not affect the speed of convergence as significantly as in other cases such as Latent Dirichlet Allocation (LDA). Hence, when training for the importance switch in each layer, we use the analytic mean of the Dirichlet random variable to make a point estimate of the integral $\int q_{\vphi_l}(\vs_l) \log p(\Dat|\vs_l) d\vs_l \approx \log p(\Dat|\tilde{\vs}_l) $, where $\tilde{\vs}_{l,j} = \vphi_{l,j}/\sum_{j'=1}^{D_l} \vphi_{l,j'}$, which allows us to directly compute the gradient of the quantity without sampling from the posterior.  As illustrated in \secref{Experiments},  this approximation performs well with relatively low dimensional switches. 


\section{Experiments}\label{sec:Experiments}
 
 In this section we present experimental results based on the two proposed approaches for CNN features ranking, the Shapley value and the importance switch methods. The tests have been performed on LeNet-5 trained on MNIST and FashionMNIST, and VGG-16 trained on CIFAR-10.

To compute the rankings for both methods the same pretrained model is used.
To compute the Shapley value of each neuron in the trained model, 
%
%
we remove the subsets of features (both weights and biases) and test the network on a validation set. As mentioned, the accuracy is the payoff for a given group of features. The computation of the complete set of payoffs is of combinatorial nature and therefore we compute the power set for layers up to 25 nodes. To account for this limitation and to illustrate better the proposed method, we choose to limit the number of nodes in the pretrained LeNet-5 architecture to 10-20-100-25. When using the trained VGG-16, we use the same number of filters in each layer as in the original architecture. For the layers with larger number of features, we use one of the two methods to compute marginal contributions. The first method uses \eqref{permutations} and only limits the number of coalitions we consider to compute SV. The second method uses \eqref{sets}  the accuracy change between two subsets which differ by a single node. Both node and the first combination were sampled uniformly at random.

  When we learn the importance switches, we load the same train model which has been used to compute the Shapley value and then only add parameters for switches and trained them per layer with fixing all the other network parameters to the trained values. We run the training of the importance switches for 300 epochs, however, in practice, even a few iterations is sufficient to distinguish important nodes from the rest.
  
  
  \paragraph{Method comparison:} We start with comparing the learnt ranks of the two methods. As summarized in Table 1, the first striking observation is that for the model pretrained both on MNIST and FashionMNIST both methods have identified similar nodes to be the most important. The similarity is more significant for smaller layers where over 50\% of top nodes (here we consider top-5 nodes for clarity and top-10 nodes for the large fc1 layer) and in three out of six cases the top two nodes are the same. Significantly for conv2 on MNIST the group of four nodes are the same, and as far as fc2 on FashionMNIST is concerned, the top five nodes chosen from the set of 25 nodes are the same (the probability to select this subset at random is $6 \cdot 10^{-5}$), showing that the methods agree when it comes to both convolutional and fully connected layers. For brevity, please look at the Appendix for the rankings of the less significant nodes but what is notable is that both methods also identified similar groups of unimportant nodes, particularly in fc2 where every node indexed higher than 9 (as compared to nodes indexed lower than 9) scored very low for both methods. When it comes to larger layers, the methods however are more discrepant (yet still significantly the common nodes are found as seen in the case of fc1 layer).  The differences may also come from the inexact computation of the Shapley value.

  \begin{center}
\begin{tabular}{|c|c|c|c|}
\hline
\textbf{Layer} & \textbf{Alg} & \textbf{FashionMNIST}  & \textbf{MNIST}  \\ \hline
   
   conv1 & SH & \textbf{0},\textbf{7},6,5,1 & \textbf{1},\textbf{8},7,4,\textbf{6} \\
   
   (10) & IS & \textbf{0},\textbf{7},5,9,6 & \textbf{8},\textbf{1},3,9,\textbf{6} \\  \hline
   
     conv2 & SH & \textbf{5},10,0,\textbf{13},9 & \textbf{2},\textbf{8},\textbf{9},\textbf{19},4\\
   (20) & IS & \textbf{5},8,\textbf{13},14,15 & \textbf{9},\textbf{2},\textbf{8},\textbf{19},6 \\  \hline
   
     fc1 & SH & \textbf{60}, \textbf{13}, 43, 88, 94, 20, 70, 44, 32, 64  & \textbf{56}, 86, 25, 64, 33, 17, \textbf{23}, \textbf{96}, 52, 81  \\
   (100) & IS & 94, 7, 50, 92, \textbf{13}, \textbf{25}, \textbf{60}, 40, 75, 45 & 25, \textbf{96}, 58, \textbf{56}, 88, 52, \textbf{23}, 43, 30, 4 \\  \hline
   
     fc2 & SH & \textbf{5},\textbf{1},\textbf{8},\textbf{9},\textbf{7} & \textbf{1},\textbf{7},2,3,0 \\
    (25) & IS & \textbf{1},\textbf{7},\textbf{9},\textbf{5},\textbf{8} & \textbf{7},\textbf{1},4,6,9 \\  \hline

\end{tabular}
\captionof{table}{Rankings of filters for the Shapley value (SH) and the importance switches (IS) methods on a four-layer network, 10-20-100-25. For each layer the top five neurons are shown, the numbers in bold indicate the common top neurons across both the methods.} 
\end{center}
\paragraph{Interpretability:} One of the main aims of this work has been to understand better the process of learning of convolutional neural networks. Building on the previous works which visualized CNN filters, we want to add an extra component and interpret that visual features by means of the filter rankings. We visualize feature maps produced by the first convolution layer of filters. Knowing the important filters allows to ponder over what features the network learns and deems useful. Given the visual results, we may make the following observations. As seen on the MNIST digits, the learnt filters identify local parts of the image (such as lower and upper parts of the digit '2' and opposite parts of the digit '0'). The interesting observation is that the most important features, on the one hand, complement each other (such as complementing parts of the digit '0' or the dog in CIFAR-10) but, on the other and, overlap to seemingly reinforce its importance. Finally, the important features appear smoother as compared to unimportant ones, which outline the object with no particular focus.

\begin{figure}[!tbp]
  \centering
  %
  %
  \begin{minipage}[b]{0.085\textwidth}
    \includegraphics[width=\textwidth]{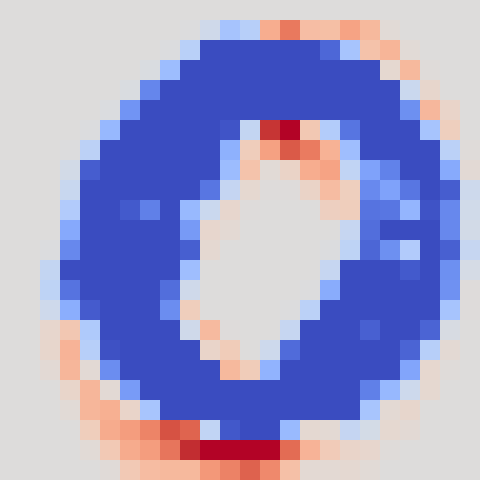}
  \end{minipage}
  \begin{minipage}[b]{0.085\textwidth}
    \includegraphics[width=\textwidth]{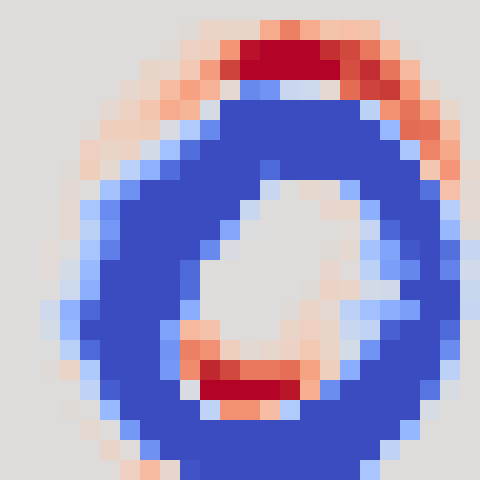}
  \end{minipage}
  \begin{minipage}[b]{0.085\textwidth}
    \includegraphics[width=\textwidth]{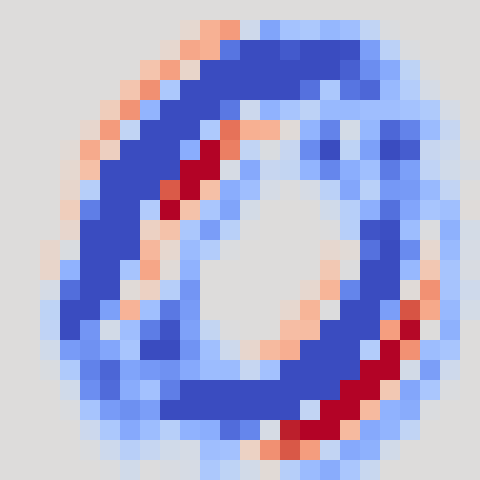}
  \end{minipage}
  \begin{minipage}[b]{0.085\textwidth}
    \includegraphics[width=\textwidth]{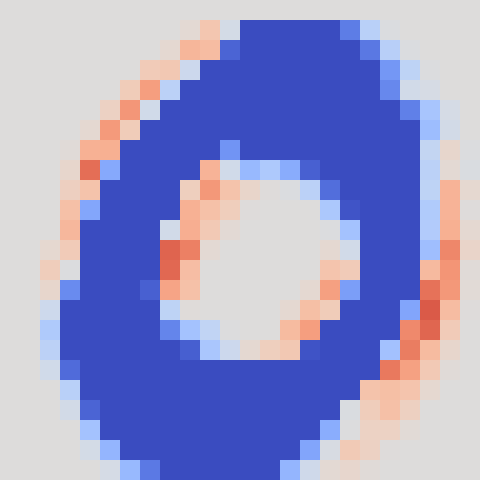}
  \end{minipage}
  \hfill
   \begin{minipage}[b]{0.085\textwidth}
    \includegraphics[width=\textwidth]{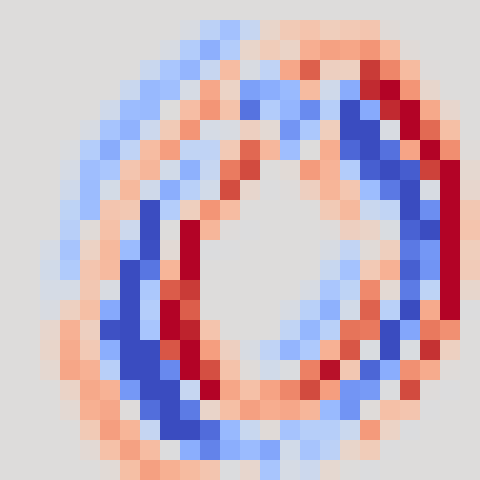}
  \end{minipage}
  \hfill
  %
  %
    \begin{minipage}[b]{0.085\textwidth}
    \includegraphics[width=\textwidth]{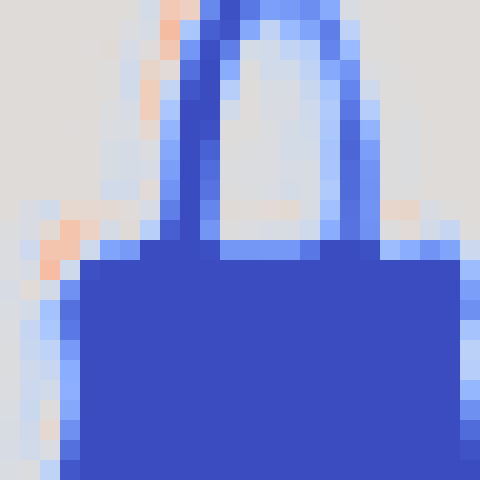}
  \end{minipage}
  \begin{minipage}[b]{0.085\textwidth}
    \includegraphics[width=\textwidth]{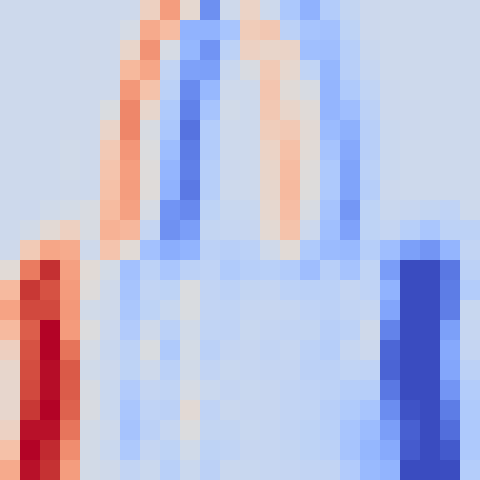}
  \end{minipage}
  \begin{minipage}[b]{0.085\textwidth}
    \includegraphics[width=\textwidth]{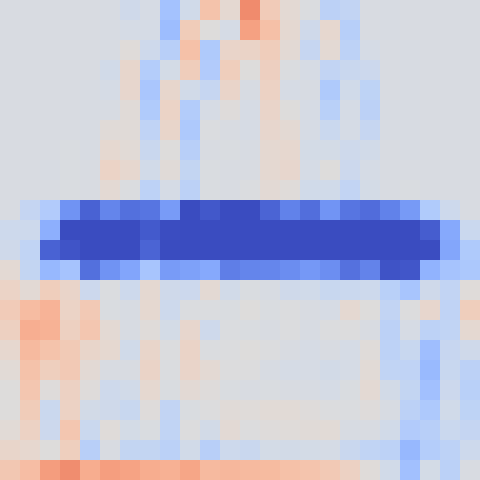}
  \end{minipage}
   \begin{minipage}[b]{0.085\textwidth}
    \includegraphics[width=\textwidth]{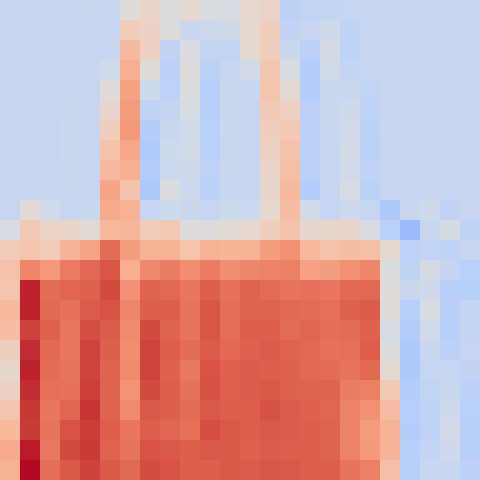}
  \end{minipage}\hfill
   \begin{minipage}[b]{0.085\textwidth}
    \includegraphics[width=\textwidth]{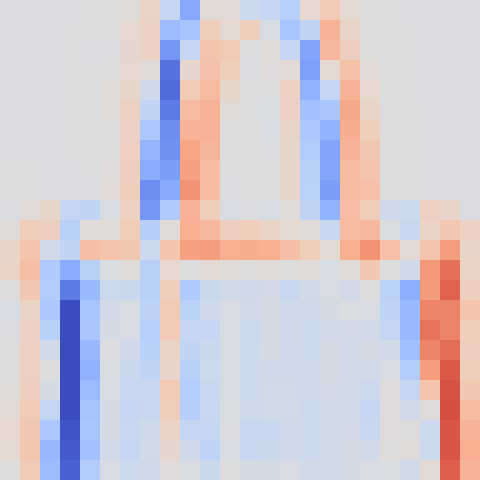}
  \end{minipage}
  
  %
  %
    \begin{minipage}[b]{0.085\textwidth}
    \includegraphics[width=\textwidth]{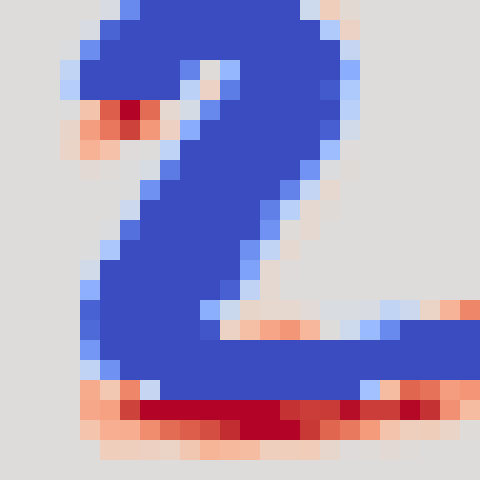}
  \end{minipage}
  \begin{minipage}[b]{0.085\textwidth}
    \includegraphics[width=\textwidth]{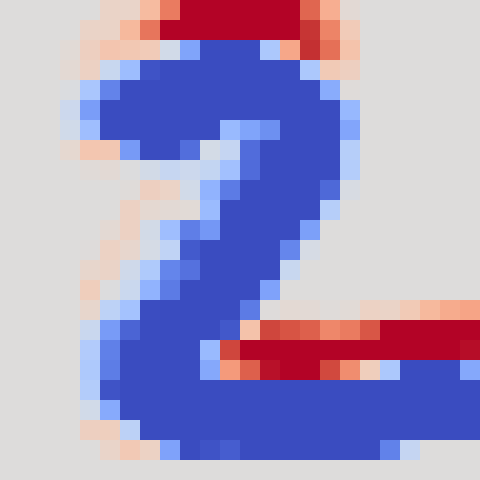}
  \end{minipage}
  \begin{minipage}[b]{0.085\textwidth}
    \includegraphics[width=\textwidth]{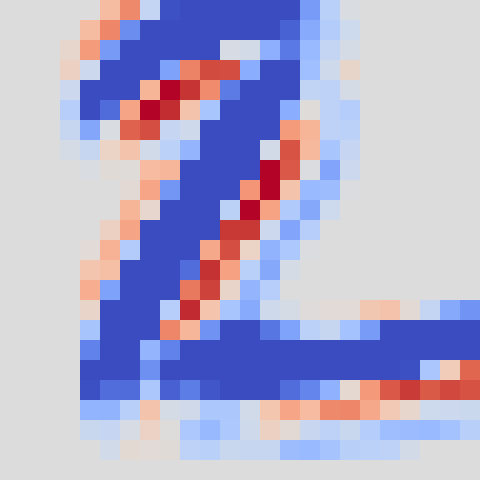}
  \end{minipage}
  \begin{minipage}[b]{0.085\textwidth}
    \includegraphics[width=\textwidth]{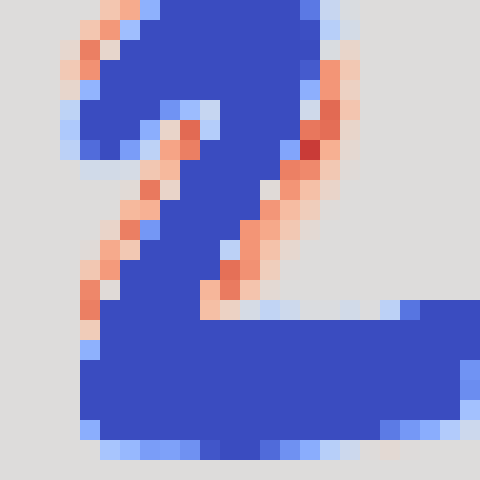}
  \end{minipage}
  \hfill
   \begin{minipage}[b]{0.085\textwidth}
    \includegraphics[width=\textwidth]{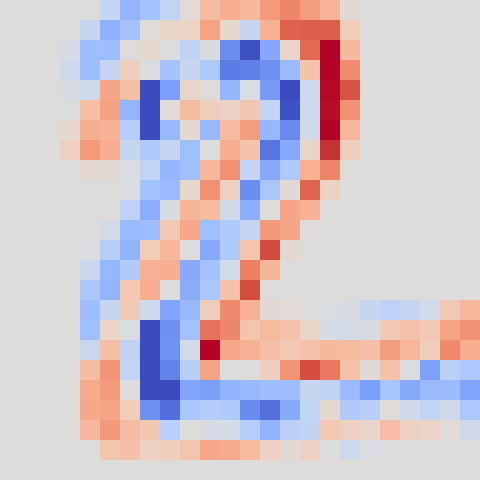}
  \end{minipage}
  \hfill
  %
  %
    \begin{minipage}[b]{0.085\textwidth}
    \includegraphics[width=\textwidth]{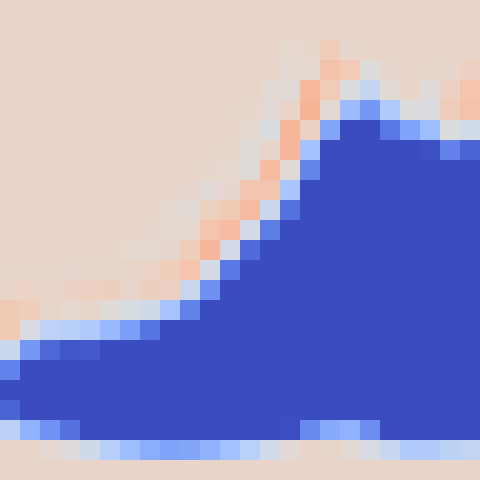}
  \end{minipage}
  \begin{minipage}[b]{0.085\textwidth}
    \includegraphics[width=\textwidth]{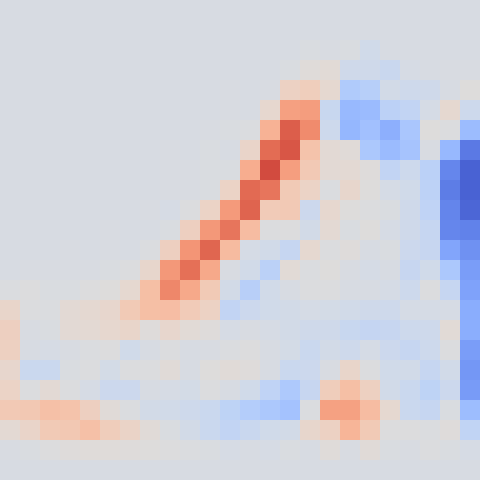}
  \end{minipage}
  \begin{minipage}[b]{0.085\textwidth}
    \includegraphics[width=\textwidth]{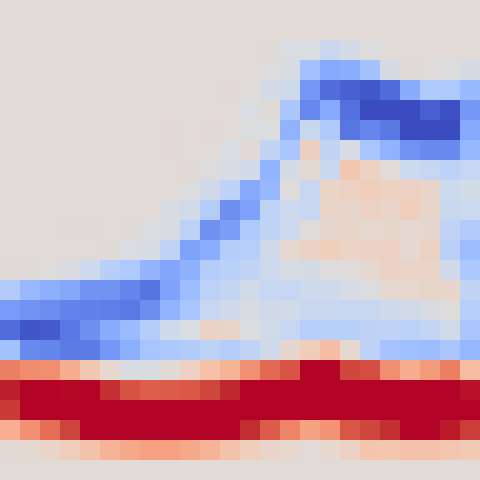}
  \end{minipage}
   \begin{minipage}[b]{0.085\textwidth}
    \includegraphics[width=\textwidth]{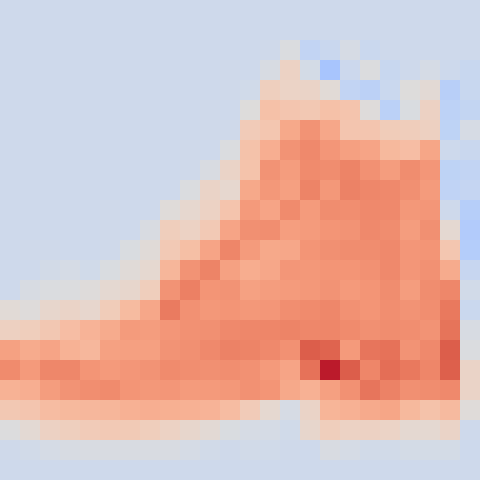}
  \end{minipage}\hfill
   \begin{minipage}[b]{0.085\textwidth}
    \includegraphics[width=\textwidth]{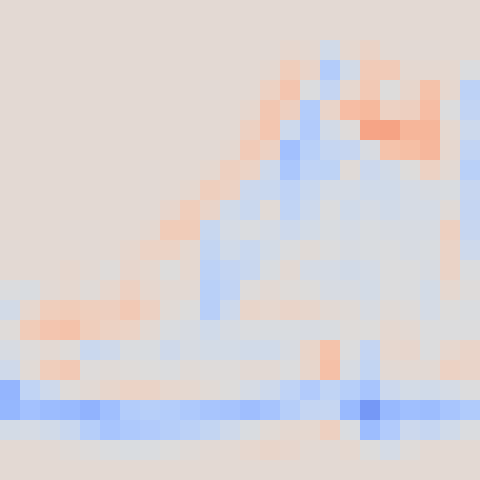}
  \end{minipage}
  
      \begin{minipage}[b]{0.085\textwidth}
    \includegraphics[width=\textwidth]{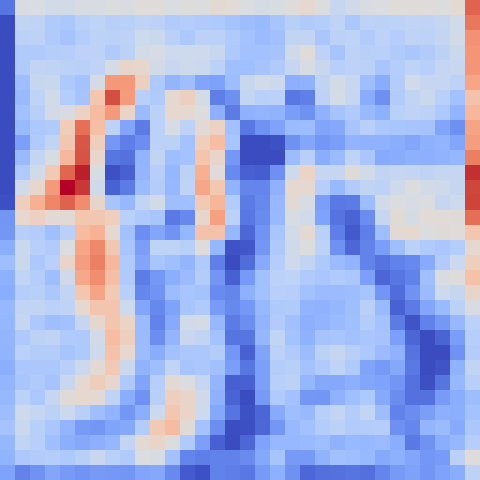}
  \end{minipage}
  \begin{minipage}[b]{0.085\textwidth}
    \includegraphics[width=\textwidth]{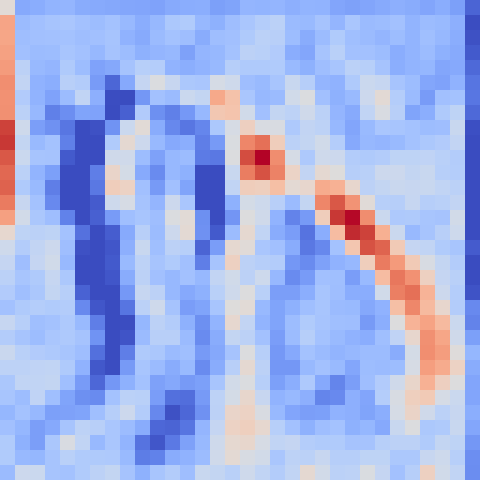}
  \end{minipage}
  \begin{minipage}[b]{0.085\textwidth}
    \includegraphics[width=\textwidth]{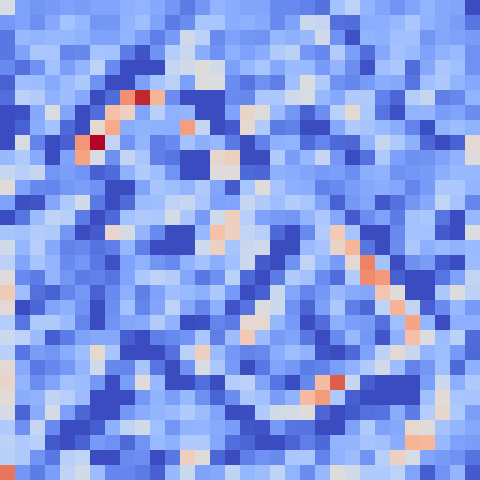}
  \end{minipage}
  \begin{minipage}[b]{0.085\textwidth}
    \includegraphics[width=\textwidth]{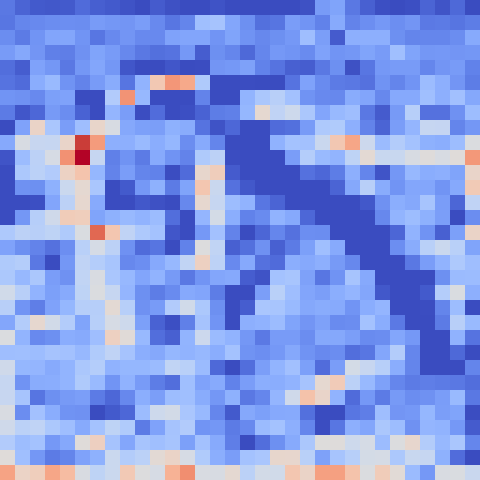}
  \end{minipage}
  \hfill
   \begin{minipage}[b]{0.085\textwidth}
    \includegraphics[width=\textwidth]{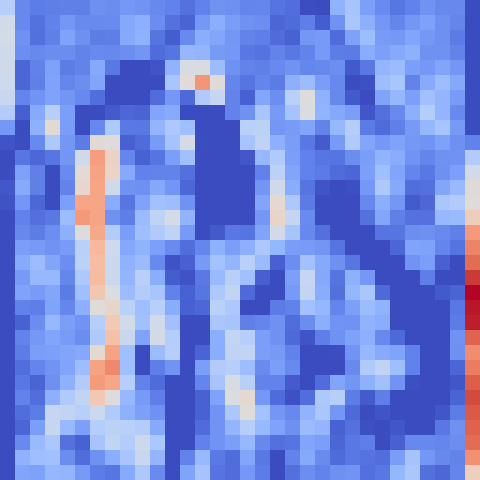}
  \end{minipage}
  \hfill
  %
  %
    \begin{minipage}[b]{0.085\textwidth}
    \includegraphics[width=\textwidth]{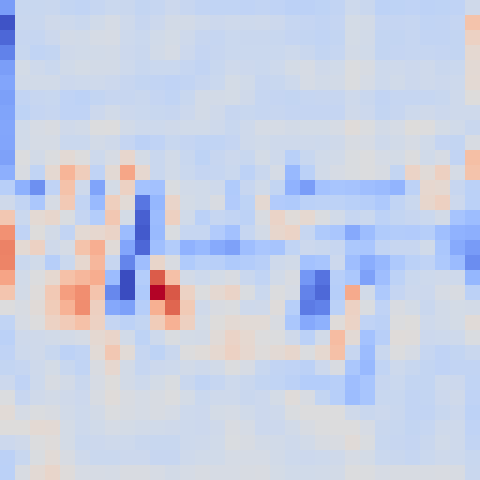}
  \end{minipage}
  \begin{minipage}[b]{0.085\textwidth}
    \includegraphics[width=\textwidth]{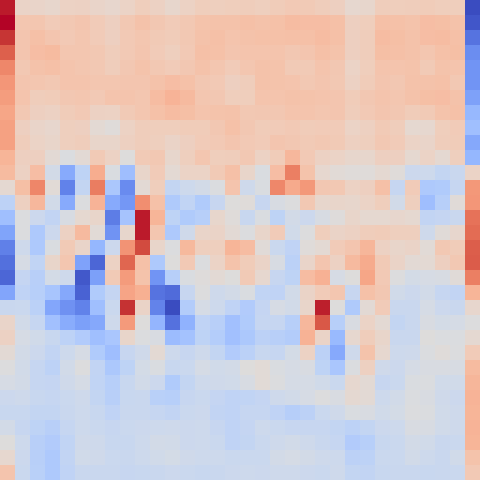}
  \end{minipage}
  \begin{minipage}[b]{0.085\textwidth}
    \includegraphics[width=\textwidth]{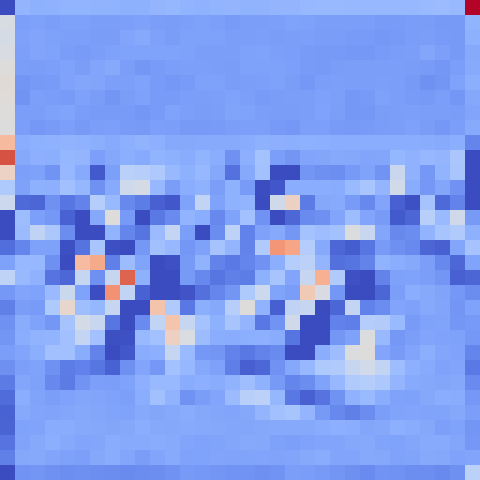}
  \end{minipage}
   \begin{minipage}[b]{0.085\textwidth}
    \includegraphics[width=\textwidth]{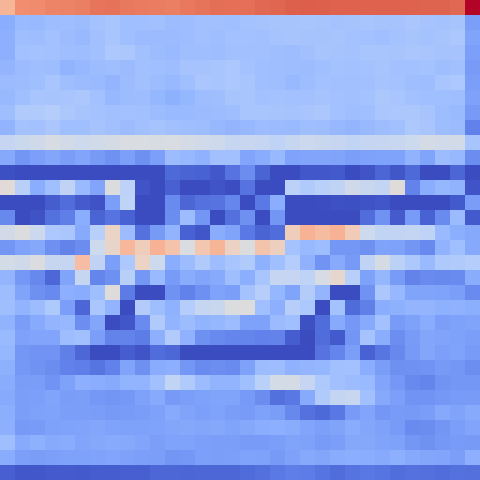}
  \end{minipage}\hfill
   \begin{minipage}[b]{0.085\textwidth}
    \includegraphics[width=\textwidth]{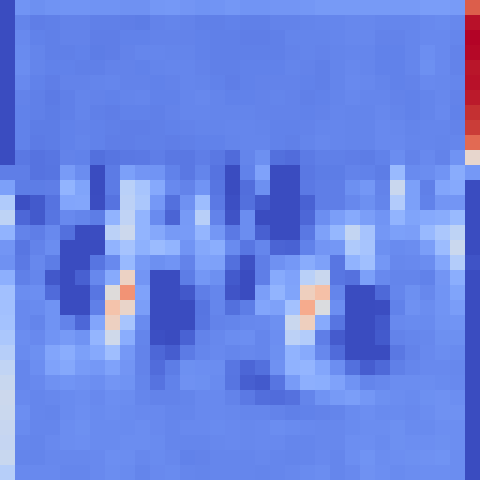}
  \end{minipage}

  \caption{Visualization of four important feature maps (MNIST: 1,8,3,7, FashionMNIST: 0,7,5,6) and one unimportant one for two examples of digits for the same filters. The third row depicts the feature maps from CIFAR-10. Notice the complementary nature of important features on MNIST: (1,3), FashionMNIST (0,7) and CIFAR-10 (the first two) and reinforcing features: MNIST (1,8), FashionMNIST (7,5). The unimportant features are jagged or lack concrete focus.}
\end{figure}


 \vspace{1cm}
\paragraph{Compression:} The consequence of the feature ranking is that some of the nodes within each layer are less significant than others and, as argued in network compression literature, the network may do as well without them. The compression experiments procedure follows from the previous experiments. Given the rankings, we prune the neurons from the bottom of the ranking and then we retrain the network. We run the tests for both of the methods on several different architectures. In all the trainings we use SGD with decreasing learning rate from 0.1 to 0.001, momentum, 0.9, weight decay, 5e-4, and early stopping. 

Table 2 presents the results for LeNet-5 as trained on MNIST, and VGG-16 as trained on CIFAR-10. For LeNet-5, the compressed architecture has 17K parameters which is less than all the other methods, and 137K FLOPs which is second to FDOO(100) \citep{tang2018flops}, which however has over three times more parameters. The method fares relatively well also on VGG producing an architecture which is smaller than others in the earlier layers but larger in later layers (the second proposed architecture has overall the least number of parameters at the cost of the performance, though). We hope to test a larger set of possible architectures in the future and devise a way to combine both rankings for a more optimal compression.
Nevertheless, the results show that the neuron ranking method is adequate for condensing both small and large architectures.


%
%
 


\begin{table}[]
\label{tab:table_comp}
    \centering
    \begin{tabular}{|c|c|c|}
    \hline
        Method & Architecture & Error\\
  	\hline
          \textbf{NR (proposed)} &\textbf{5-7-45-20}&  1.0\% \\
        
          BC-GNJ & 8-13-88-13&  1.0\%\\
          BC-GHS & 5-10-76-16 & 1.0\%\\
          FDOO(100K) & 2-7-112-478&  1.1\%\\
          FDOO(200K) & 3-8-128-499& 1.0\% \\
        
           GL & 3-12-192-500&  1.0\% \\
           GD & 7-13-208-16&  1.1\% \\
            SBP & 3-18-284-283& 0.9\% \\
        \hline
        
         
           \textbf{NR (proposed)} & \textbf{34-34-68-68-75-106-101-92-102-92-92-67-67-62-62}  & 8.6\%\\
        
        & \textbf{39-39-63-48-55-98-97-52-62-22-42-47-47-42-62}  & 9.1\%\\ 
        
         BC-GNJ & 63-64-128-128-245-155-63-26-24-20-14-12-11-11-15& 8.3\% \\

          BC-GHS & 51-62-125-128-228-129-38-13-9-6-5-6-6-6-20 & 8.3\%\\
          \hline
    
    \end{tabular}
    \vspace{0.5cm}
    \caption{The structured pruning of LeNet-5 on MNIST. We benchmark our method against BC-GNJ, Bayesian Compression - Group Horseshoe  (BC-GHS)~\citep{louizos2017bayesian}, FDOO~\citep{tang2018flops}, Generalized Dropout(GD)~\citep{srinivas2015data}, Group Lasso(GL)~\citep{wen2016learning} and 
    Structured Bayesian Pruning (SBP)~\citep{neklyudov2017structured}.
    }

\end{table}


\section{Conclusion}

In summary, this work suggests a theory that the learnable CNN features contain inherent hierarchy where some of the features are more significant than others. This multidisciplinary work which builds on top of probability and game theoretical concepts proposes two methods to produce feature ranking and select most important features in the CNN network. The striking observation is that the different methods lead to similar results and allow to distinguish important nodes with larger confidence. The ranking methods allow to build an informed way to build a slim network architecture where the significant nodes remain and unimportant nodes are discarded. A future search for further methods which allow to quantify the neuron importance is the next step to develop the understanding of the feature importance in neural networks.
\vspace{3cm}

\bibliographystyle{plainnat}
\bibliography{ref}


\newpage
\section{Appendix}

\subsection{LeNet MNIST rankings}

 \begin{figure}[H]
\begin{minipage}[b]{0.45\textwidth}
    \includegraphics[width=\textwidth]{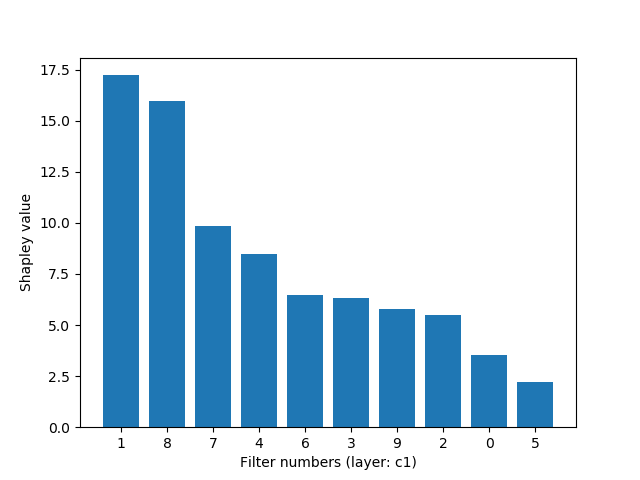}
  \end{minipage}
  \begin{minipage}[b]{0.45\textwidth}
    \includegraphics[width=\textwidth]{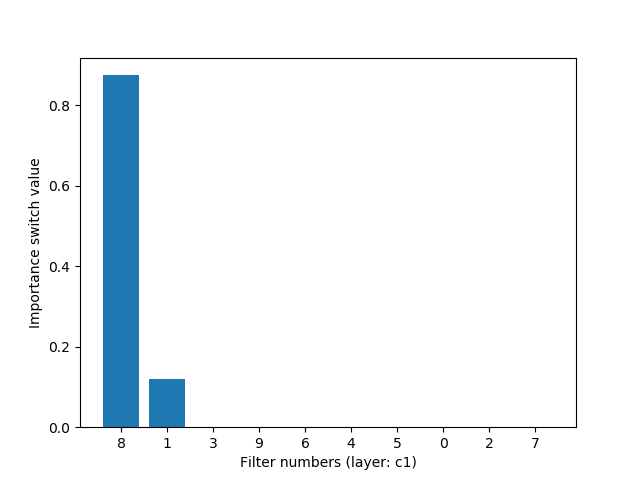}
  \end{minipage}
  \hfill
\begin{minipage}[b]{0.45\textwidth}
    \centering
    \includegraphics[width=\textwidth]{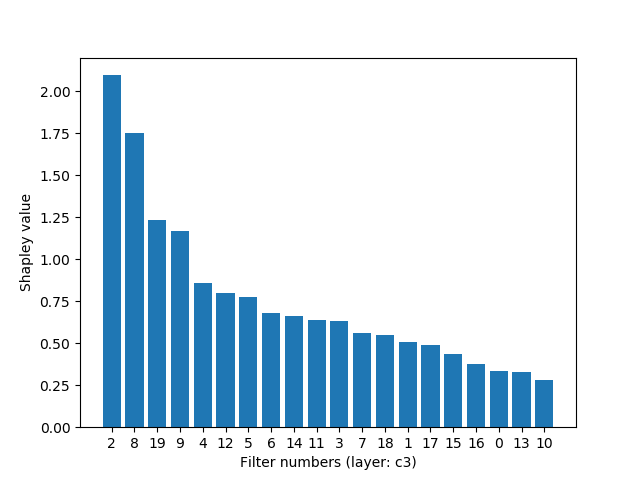}
\end{minipage}
\begin{minipage}[b]{0.45\textwidth}
    \centering
    \includegraphics[width=\textwidth]{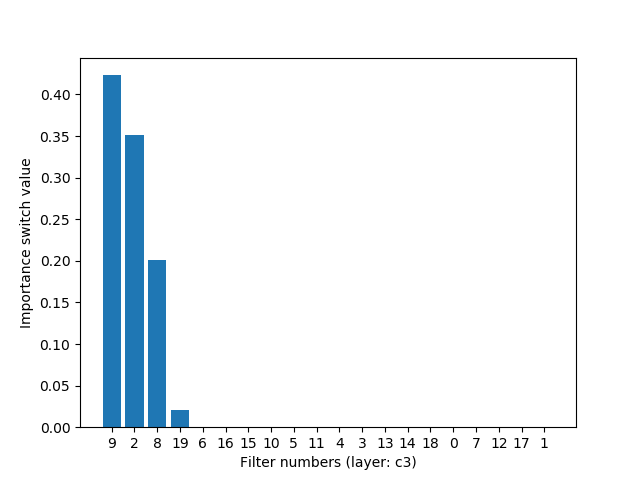}
\end{minipage}
\hfill
\begin{minipage}[b]{0.45\textwidth}
    \centering
    \includegraphics[width=\textwidth]{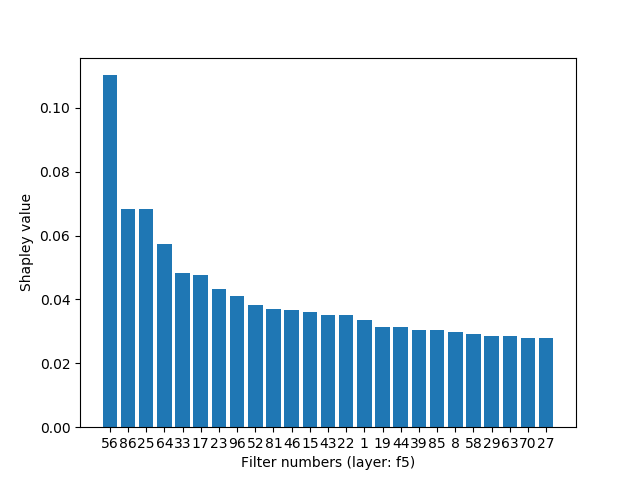}
\end{minipage}
\begin{minipage}[b]{0.45\textwidth}
    \centering
    \includegraphics[width=\textwidth]{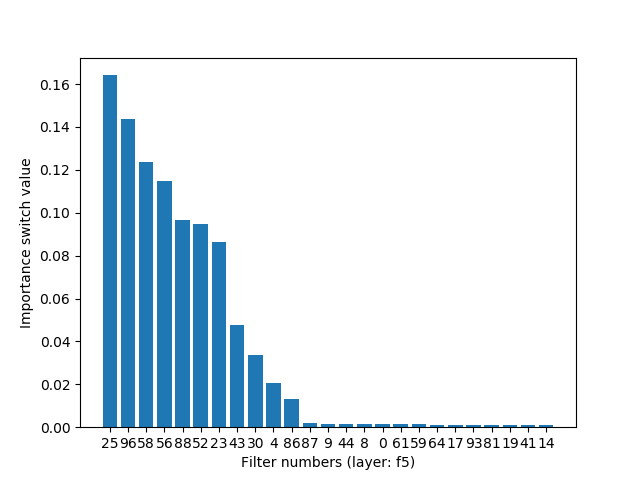}
\end{minipage}
\hfill
\begin{minipage}[b]{0.45\textwidth}
    \centering
    \includegraphics[width=\textwidth]{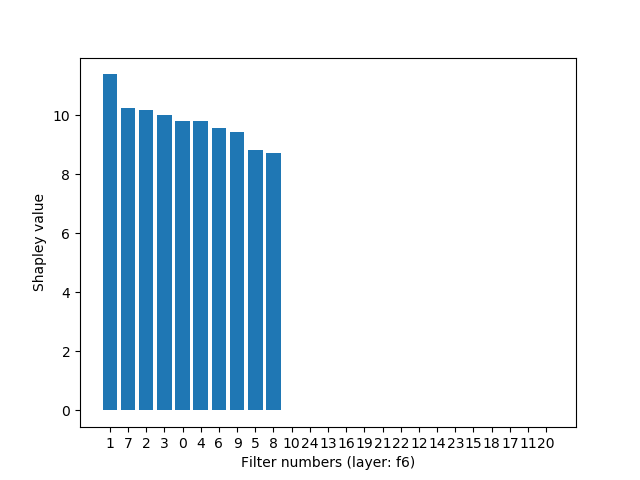}
\end{minipage}
\hspace{1.12cm}
\begin{minipage}[b]{0.45\textwidth}
    \centering
    \includegraphics[width=\textwidth]{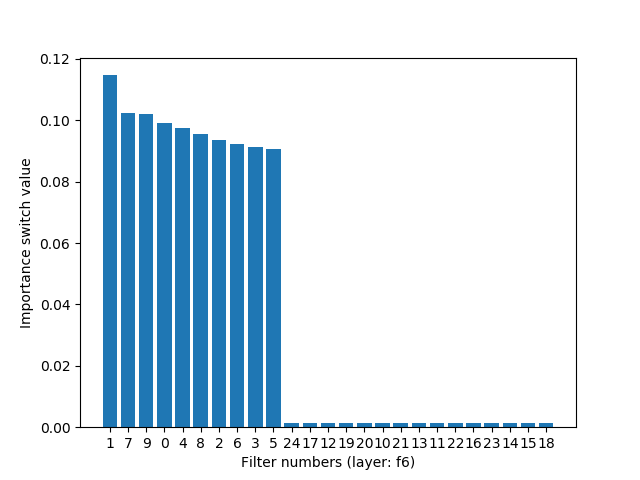}
\end{minipage}
\hfill
\captionof{figure}{The bar charts visualize filter rankingse for the LeNet network with two convolutional and two fully connected layers. The vertical axis describes, respectively, the Shapley value (left column) and the importance switches value (right column). The horizontal axis contains the filter indices.}
\end{figure}

\subsection{LeNet FashionMNIST rankings}
\begin{figure}[H]
\begin{minipage}[b]{0.45\textwidth}
    \includegraphics[width=\textwidth]{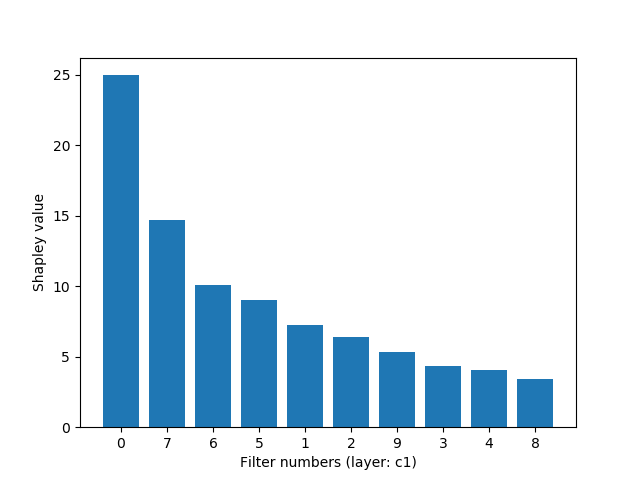}
  \end{minipage}
  \begin{minipage}[b]{0.45\textwidth}
    \includegraphics[width=\textwidth]{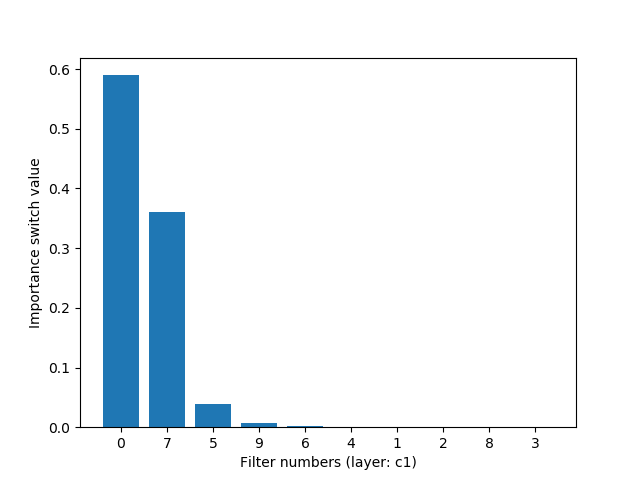}
  \end{minipage}
  \hfill
\begin{minipage}[b]{0.45\textwidth}
    \centering
    \includegraphics[width=\textwidth]{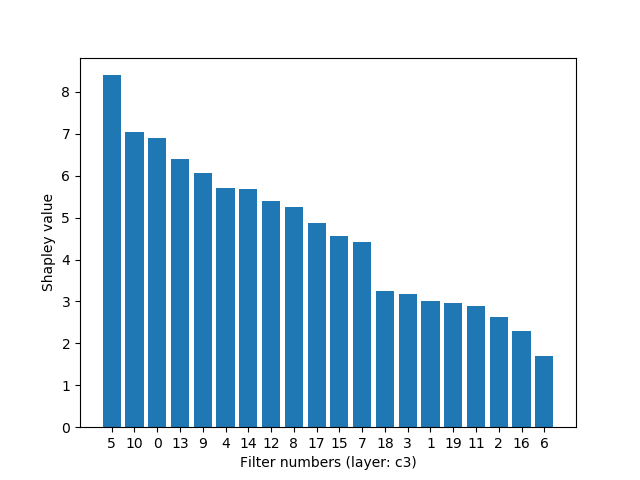}
\end{minipage}
\begin{minipage}[b]{0.45\textwidth}
    \centering
    \includegraphics[width=\textwidth]{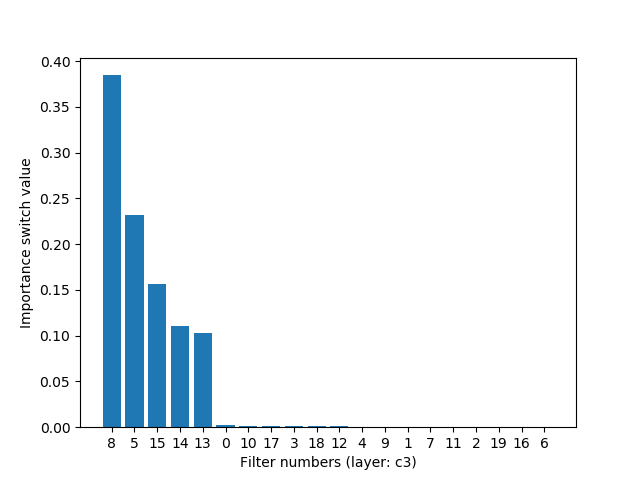}
\end{minipage}
\hfill
\begin{minipage}[b]{0.45\textwidth}
    \centering
    \includegraphics[width=\textwidth]{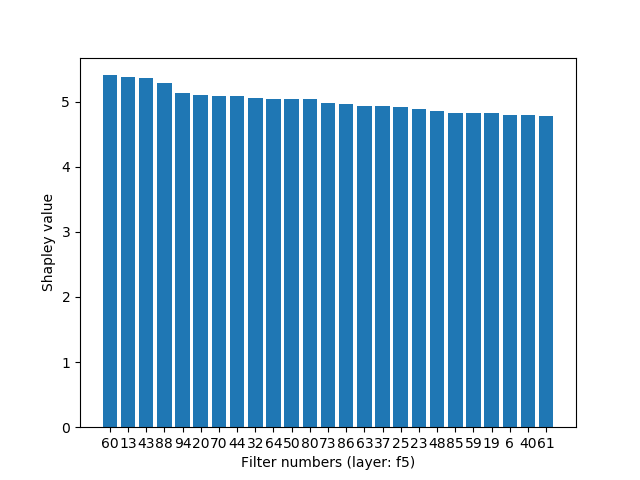}
\end{minipage}
\begin{minipage}[b]{0.45\textwidth}
    \centering
    \includegraphics[width=\textwidth]{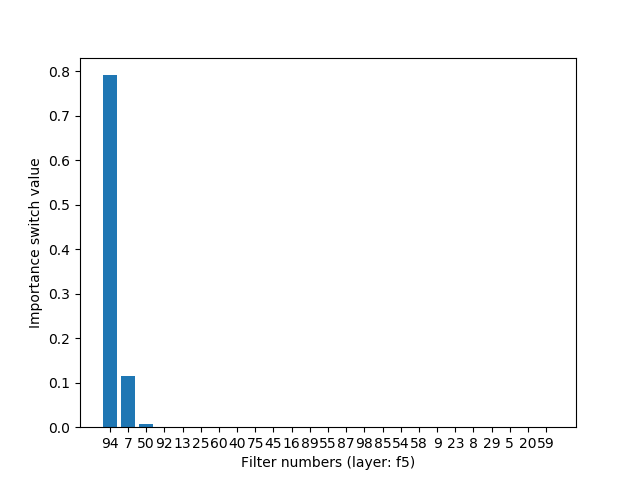}
\end{minipage}
\hfill
\begin{minipage}[b]{0.45\textwidth}
    \centering
    \includegraphics[width=\textwidth]{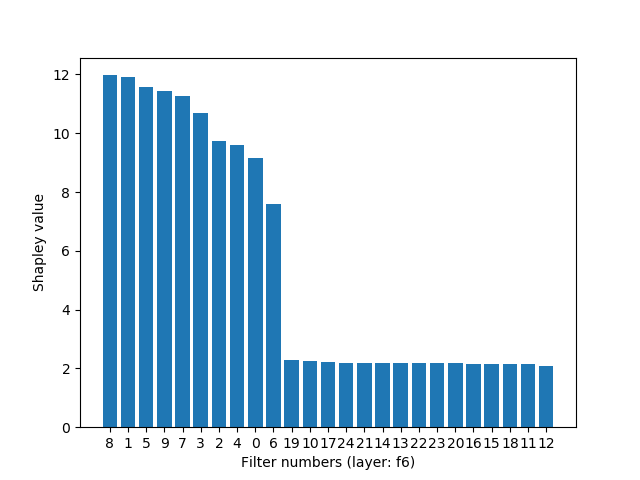}
\end{minipage}
\hspace{1.12cm}
\begin{minipage}[b]{0.45\textwidth}
    \centering
    \includegraphics[width=\textwidth]{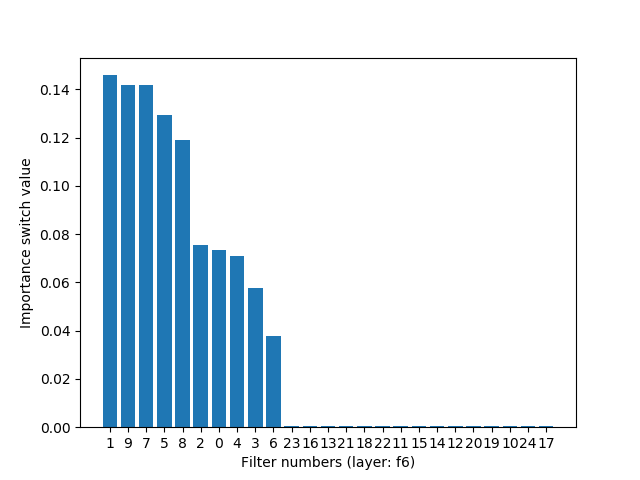}
\end{minipage}
\hfill
 \captionof{figure}{The bar charts visualize filter rankingse for the LeNet network with two convolutional and two fully connected layers. The vertical axis describes, respectively, the Shapley value (left column) and the importance switches value (right column). The horizontal axis contains the filter indices.}
 \end{figure}

\end{document}

%% file: notations.tex
\newcommand{\punt}[1]{}

\theoremstyle{remark}

\newcommand{\vx}{\mathbf{x}}
     
\newcommand{\valpha}{\bm{\alpha}}

\newcommand{\vs}{\mathbf{s}}

\newcommand{\vh}{\mathbf{h}}

\newcommand{\vz}{\mathbf{z}}

\newcommand{\vy}{\mathbf{y}}

\newcommand{\bq}{\begin{equation}}
\newcommand{\eq}{\end{equation}}
\newcommand{\ba}{\begin{eqnarray}}
\newcommand{\ea}{\end{eqnarray}}

\newcommand{\remove}[1]{}

\newcommand{\Nrm}{\mathcal{N}}

\renewcommand{\eqref}[1]{eq.~\ref{eq:#1}}

\newcommand{\secref}[1]{Sec.~\ref{sec:#1}}  

\newcommand{\Dat}{\mathcal{D}}

\newcommand{\vphi}{\mathbf{\ensuremath{\bm{\phi}}}}

\newcommand{\vone}{\mathbf{1}} 

\newcommand{\LL}{\ensuremath{\mathcal{L}}}

\newcommand{\mW}{\mathbf{W}}

\usepackage{color}  
\definecolor{gray}{rgb}{0.5, .5, .5}